\documentclass[conference]{IEEEtran}
 \usepackage{amsmath,amsthm,amssymb,amsfonts}
 \usepackage{hyperref}
 \usepackage{amssymb}
 \usepackage{array,multirow,graphicx}
 \usepackage{float}
 \usepackage{balance}
 \usepackage{comment}
\usepackage{booktabs}

 \usepackage{xcolor,soul,framed} 

\colorlet{shadecolor}{yellow}

\usepackage{array}
\usepackage{mdwmath}
\usepackage{mdwtab}
\usepackage{eqparbox}
\usepackage{url}

\title{Enhanced Review Detection and Recognition: A Platform-Agnostic Approach with Application to Online Commerce}

\begin{document}


\author{
    \IEEEauthorblockN{Priyabrata Karmakar\IEEEauthorrefmark{1}, John Hawkins\IEEEauthorrefmark{2}}
    \IEEEauthorblockA{\IEEEauthorrefmark{1}Institute of Innovation, Science and Sustainability, Federation University, Australia
    \\\{p.karmakar\}@federation.edu.au}
    \IEEEauthorblockA{\IEEEauthorrefmark{2}\href{https://transitional-ai.github.io}{Transitional AI Research Group}, Australia
    \\\{john\}@getting-data-science-done.com}
}

\maketitle
 
\begin{abstract}
Online commerce relies heavily on user generated reviews to provide unbiased information about products that they have not physically seen. The importance of reviews has attracted multiple exploitative online behaviours and requires methods for monitoring and detecting reviews. We present a machine learning methodology for review detection and extraction, and demonstrate that it generalises for use across websites that were not contained in the training data. This method promises to drive applications for automatic detection and evaluation of reviews, regardless of their source. 
Furthermore, we showcase the versatility of our method by implementing and discussing three key applications for analysing reviews: Sentiment Inconsistency Analysis, which detects and filters out unreliable reviews based on inconsistencies between ratings and comments; Multi-language support, enabling the extraction and translation of reviews from various languages without relying on HTML scraping; and Fake review detection, achieved by integrating a trained NLP model to identify and distinguish between genuine and fake reviews.

\end{abstract}


\IEEEpeerreviewmaketitle
	
\section{Introduction}

Online reviews contain evaluations and feedback provided by customers of a business. They are typically shared via the internet on various platforms such as e-commerce sites, social media, or dedicated review websites. These reviews offer insights into the quality of products and services, influencing the purchasing decisions of other potential consumers\cite{watson2022impact}.

Online reviews represent real-life experiences, offering potential customers a glimpse into the quality, value, and reliability of the product or service they are considering. It plays a crucial role in the digital age, providing social proof and influencing consumer decisions. They offer real-life experiences, giving potential customers insight into the quality, value, and reliability of products or services. Online reviews are more relatable and trustworthy than traditional advertising, contributing to informed and thought-out purchase decisions. People crave reassurance that they’re making the right decision, particularly when it comes to parting with their hard-earned money. \cite{online-reviews}.
For businesses, online reviews are essential for building trust, improving visibility, and influencing consumer decisions. They can make or break a company's success and are a key factor in creating a trusted brand. Customers are 63\% more likely to trust brands with online reviews compared to brands with no reviews. A lack of reviews makes buyers feel increased risk, which makes them less likely to buy.\cite{online-reviews2}. Additionally, online reviews can impact sales, improve a business's trustworthiness, and provide free marketing content \cite{online-reviews3}.

Although online reviews increase the amount of product information available to consumers\cite{Lackermair2013}, the advantages are not uniformly distributed\cite{Choi2020}. There are distinct problems introduced by online reviews as well, for example, it is a constant challenge for distinguishing fake reviews from the genuine ones\cite{dwivedi2021setting}. In addition, online reviews may lack context, be biased, or influenced by incentives, potentially impacting the overall credibility of the review system and the reliability of the information available to consumers \cite{otero2021fake}. A detailed discussion on advantages and limitations of online reviews is provided later in this paper.

Going through online reviews manually can be a time consuming and expensive process. Thus, there is a requirement for this process to be automated. Automatic detection enhances the scalability and timeliness of the review evaluation process, ensuring that a large number of reviews can be assessed in a relatively short time frame, especially in the context of the ever-increasing volume of online reviews \cite{WALTHER2023100278}. Furthermore, automatic detection, often leveraging natural language processing (NLP) techniques, provides a more comprehensive and objective assessment of reviews compared to manual reading, as it emphasises lexical features and textual data \cite{SALMINEN2022102771}. In the literature, many other features of websites have been automatically detected with machine learning, including logos \cite{Balan2016}.


In this paper, we proposed a novel approach to collect online review data by combining the advantages of two computer vision techniques: object detection and optical character recognition. 
Furthermore, we have also integrated three applications into our proposed approach. These applications significantly enhance the functionality and utility of our system. The first application, Sentiment Inconsistency Analysis, identifies and filters out unreliable reviews by detecting inconsistencies between sentiments expressed in ratings and comments. In addition, our system provides robust multi-language support, overcoming challenges associated with HTML scraping and enabling the extraction and translation of reviews from various languages. Lastly, we seamlessly integrate a trained NLP model for fake review detection, effectively distinguishing between genuine and fake reviews without requiring modifications to the existing pipeline.

The rest of paper is organised as follows. Sections II discusses the advantages and disadvantages of online reviews. The importance of automatic detection and recognition of online reviews is explored in Section III. The proposal is explained in Section IV. Data collection \& experiments and Results are discussed in Sections V and VI, respectively. Analysis and extended applications of the proposal is explored in Section VII. Finally Section VIII concludes the paper.


\section{Advantages and Disadvantages of Online Reviews}
In this section, we discuss the advantages and disadvantages of online reviews. 	

\subsection{Advantages}

Online reviews can reveal a lot about the businesses. A wealth of positive words can have a measurable impact on the sales, driving purchases and creating a base of consumers who will stand behind you and your product \cite{adv-online-reviews}. Some of the advantages are as follows.

\begin{enumerate}
\item{Drive Sales}: Positive reviews from other buyers boost confidence in purchasing well-reviewed items, driving sales \cite{baddeley2010herding}.

\item{Build Trust}: Good quality product and service reviews build trust, while lower ratings lead to customer distrust \cite{utz2012consumers}.

\item{Decision Making}: Online reviews provide valuable insights into product quality, functionality, and user satisfaction, helping customers make informed purchase decisions in a competitive market \cite{wang2023influence}.

\item{Enhanced Visibility}: Customer reviews contribute to search engine optimization (SEO) by including business names and product/service keywords, increasing organic traffic \cite{dwivedi2021setting}.

\item{Proactive Customer Engagement and Continuous Improvement}: Addressing negative feedback demonstrates commitment to customer satisfaction, fosters positive relationships, and showcases dedication to continuous improvement and customer service excellence \cite{chen2022impact}.

\end{enumerate}

\subsection{Disadvantages}

Although there are huge benefits of online reviews, there are many limitations associated with it as well. In this section, we discuss some of the disadvantages of online reviews.

\begin{enumerate}
\item{Fake Reviews}: Deceptive reviews, created by competitors or malicious individuals, distort ratings and mislead consumers, posing a challenge to review platforms' integrity \cite{wu2020fake}.

\item{Reputation and Revenue}: Negative online reviews can swiftly damage a business's reputation and deter potential customers, leading to revenue loss \cite{otero2021fake}.

\item{SEO Ranking}: Negative reviews can significantly impact a business's search engine ranking, undermining SEO efforts \cite{online-reviews-seo}.

\item{Cost of Third-Party Reviews}: Using third-party review sites often incurs additional costs, impacting budget and resources \cite{choi2023fake}.
\end{enumerate}

\section{Importance of Automatic Detection and Recognition of Online Reviews}

As the advantages and disadvantages of online reviews are discussed in the above section, it can be clearly understood that how important the online reviews for any business. Therefore, it is essential that online reviews are detected and recognised efficiently and effectively. So that the corresponding businesses can take necessary actions accordingly. In addition, it is also essential for online market places like Amazon, eBay to detect and identify fake reviews and remove them in order to maintain their platform's integrity. Going through online reviews and assess them manually by humans is an expensive and time consuming process. This can be addressed by the implementation of automatic detection and recognition of online reviews and it has many benefits. They are as follows \cite{WALTHER2023100278}.

\begin{enumerate}
    \item {Volume and Scalability}: The volume of online reviews can be massive, impacting consumers' understanding of product information and, consequently, their purchase decisions. Automatic detection enables the efficient processing of a large number of reviews, which is challenging to achieve manually.

    \item{Increased Efficiency}: Automatic detection allows for timely identification of fake reviews, which can significantly impact consumer trust and purchase decisions. Manual review analysis is time-consuming and may not keep pace with the rapidly increasing volume of online reviews \cite{deshai2023unmasking}.

    \item{Automated Review Insights}: Automatic detection, often leveraging machine learning and natural language processing, provides a more comprehensive and objective assessment of reviews compared to manual review reading. This can help in identifying patterns, sentiments and trends that may not be readily apparent through manual analysis \cite{SALMINEN2022102771, kang2022study}.

\end{enumerate}

The popular approach to extract online reviews is mostly through web scrapping that involves making HTTP requests to a website's server to retrieve the HTML or XML source code of a web page, allowing businesses to extract large amounts of data quickly and efficiently \cite{jishag2019automated, guyt2024unlocking}. However, web scrapping is unable to retrieve data from all websites, particularly e-commerce websites, at an affordable cost. In addition, each website is unique, with its own pattern and programming logic, making it challenging to extract data from every site \cite{web-scrapping}. Therefore, there is a need for a robust framework  for the detection and recognition of online reviews.

\section{Methods}

\begin{figure*}
    \centering
    \includegraphics[width=1\linewidth]{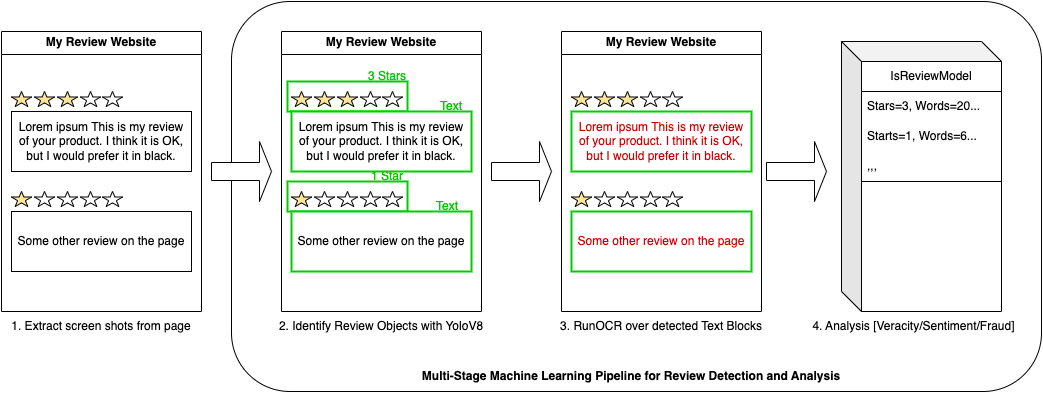}
    \caption{Proposed Multi-Stage Machine Learning Process for Automatic Review Detection and Analysis}
    \label{fig:process}
\end{figure*}

In this paper, we propose a novel approach to detecting online reviews using computer vision (CV) for automatic detection on arbitrary websites. Specifically, we use object detection to detect the sections relevant to the reviews on a web page followed by applying optical character recognition (OCR) to recognise the text in the detected reviews. This general approach is depicted schematically in Figure \ref{fig:process}.

We argue this approach is a more robust method compared to web scraping for the following reasons:
\begin{itemize}
    \item Web scraping relies on predefined HTML tags, and changes in these tags can lead to failures in extracting desired information \cite{afonso2024development}. Object detection and OCR on the other hand rely on the visual presentation regardless of the underlying HTML.
    \item Arrangement of reviews is often demarcated by consistent visual elements that can be implemented in many different ways in HTML but will be visually similar.
    \item Object detection and OCR can understand the semantic meaning of the content of the underlying document. Therefore, they are highly effective in detecting and recognise objects and texts. Web scrapping lacks this characteristic.
\end{itemize}

\begin{figure}
    \centering
    \includegraphics[width=1\linewidth]{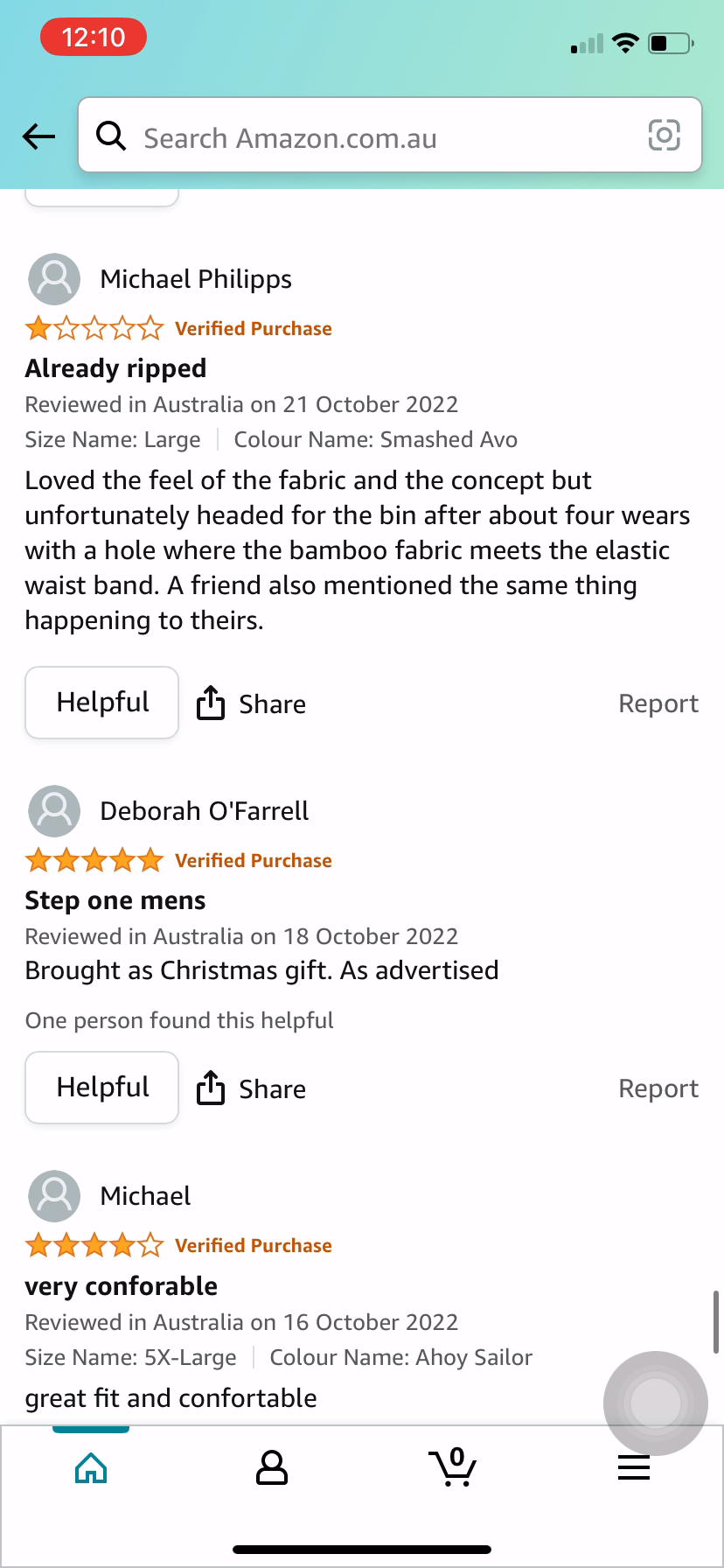}
    \caption{Sample screenshot of online reviews from Amazon.com.au platform}
    \label{fig:sample-amazon}
\end{figure}

For object detection, we have adopted the Yolov8 model \cite{yolov8} trained on our custom data. Pytesseract \cite{pytesseract} is applied to recognise the review texts that are detected by the trained Yolov8 model. In this context, one question may arise regarding the need of using both object detection and OCR when OCR itself can scan an image and output the texts in it. The reason for this is two-fold:

\begin{enumerate}
   
 \item  In case only OCR is used to scan the image where reviews exist, it will return the texts of the entire image without providing any information whether a text belong to actually a user's review or the text is related to username of the reviewer or the date-time when the review was written. Also, an image may contain more than one review, out of which a review can be positive and another can be negative. Therefore, it would be difficult to classify a text belong to which specific review.
 \item  OCR by default scans the entire image. It detects potential text areas and then recognises the texts in the detected areas. Therefore, OCR outputs any text available in that image irrespective of it's relevance. Thus, it causes increased time complexity due to the detection and recognition of irrelevant texts. For example, Fig \ref{fig:sample-amazon} shows a sample screenshot of online reviews from Amazon Australia platform. In this screenshot, there are some irrelevant texts exist like Report or the texts at Helpful or Share button. The count of these irrelevant texts may be less compared to the relevant texts but when bulk of images are processed, the detection and recognition of these irrelevant texts will degrade the overall efficiency.

\end{enumerate}

In the proposed approach, we aim to address the above limitations by integrating object detection and the recognition function of OCR. We have trained a custom YOLOv8 model with custom data, comprising annotated screenshots of online reviews. During inference, the model detects only the specified areas of interest (AOI) in the screenshot. Subsequently, Pytesseract is used to recognise text only in the detected AOIs. This approach reduces the time complexity in two ways: (a) by decreasing the detection time, as only the AOIs need to be detected, and (b) by reducing the recognition time, as only the texts in the AOIs need to be recognised. Moreover, by detecting AOIs first, it ensures that the recognised texts belong to a single review, simplifying the analysis using NLP techniques. Furthermore, our custom YOLOv8 model was trained to identify star ratings, when relevant. By using the bounding box coordinates of detected objects, star ratings can easily be associated with the specific reviews. This task may pose challenges if relying only on OCR. In addition, the presence of both star rating objects and text blocks on a page can be used as additional filtering criteria when determining whether a block of text belongs to a genuine review. The detection of multiple elements in a specific geometric arrangement on the page can form the basis of adaptive post-processing logic.


We argue that on most review platforms, the elements of the reviews themselves are visually similar. This is in large part because e-commerce retailers want their customers to recognise that the page contains reviews which can influence their purchasing\cite{gulfraz2022understanding}. Therefore, it is our hypothesis that the object detection model trained on the data from several review platforms will be able to detect review elements from the screenshots of other review platforms. Our proposed approach is a model for platform-independent online reviews detection and recognition (PIORDR). The resulting model will not require training a custom object detection model for each review platform it is applied to. Instead, PIORDR can be used for most existing review platforms as well as any new platforms in future. We will explore this possibility in the experiments and results section.

\section{Data collection \& Experiments}

For training the YOLOv8 object detection model we created a custom dataset, collecting data from two major mobile application platforms. They are: Amazon (background: light) and Apple app store (background: dark). At first, we screen recorded a user scrolling through the the review sections of these two platforms on a mobile device. The frames were then extracted from the recordings at 1 frames per second (i.e., FPS = 1). We then annotated the frames as per defined AOI using LabelImg annotation tool \cite{tzutalin2015tzutalin}. The total numbers of frames and labeled reviews are shown in Table \ref{tab:train_data}.

\begin{table}
\centering
\caption{Data Summary}
\label{tab:train_data}
\begin{tabular}{l||l|r|r}
\toprule
Set      &Source Website         &Frames      &Reviews \\
\midrule
Training &Amazon         &  400        & 1255   \\
         &Apple App Store        &  380        & 1006   \\   
\midrule
Testing  &Amazon          &70          &216      \\
         &Apple App Store        &67          &172      \\  
         &Menulog           &25          &60      \\  
         &Product Review         &30          &58      \\
         &Booking.com            &26          &55      \\ 
         &eBay            &20          &65      \\ 
\bottomrule
\end{tabular}
\end{table}

In addition, we collected test data from four additional review sources to evaluate the performance of our approach on novel websites. These additional sources and the data volumes are shown in Table \ref{tab:train_data}.

To initialise the experiment we started with the default weights of YOLOv8, pre-trained on COCO dataset\cite{Lin2014}. The training is performed on a CUDA enabled Tesla T4 GPU for a total of 300 epochs. We trained the model for six different labels. One label for each ratings (i.e., 1 to 5 start ratings, so total five labels) and the sixth label represents the review text. The labels related to ratings will be applicable to only the platforms that allow a rating system. However, review text is applicable to any online reviews platform. 

\subsection{Metrics}

We evaluate the performance of our models on the test data using two metrics, the Precision (P) \cite{hawkins2021minvime} of review texts given by \eqref{eq:precision} and the mean Average Precision (mAP) \cite{karmakar2021novel} given by \eqref{eq:mAP}. The reason for separating the precision on review text objects independently is that extraction of review text content is the primary application focus.

\begin{equation}
\label{eq:precision}
    Precision = \frac{True Positive (TP) }{True Positive (TP) + False Positive (FP)},
\end{equation} the TP and FP values of review text detection is calculated by considering both thresholds of detection confidence and intersection over union (IoU) as 0.8.

The Average Precision (AP) for object detection models is calculated over a range of thresholds in the IoU. The IoU is a metric that describes the difference between the labelled boundary in the test data and the detected boundary by the model. We follow the standard practice of taking thresholds from 0.5 to 0.95 with a step size of 0.05.

\begin{equation}
\label{eq:mAP}
mAP = \frac{1}{N} \sum_{i=1}^{N} AP_i,
\end{equation} where $N$ is the total number of classes.

The mAP metric is computed by first taking the average precision for each object class the model is trained to detect. We then take the mean of these average precisions over all of the object classes the model is trained to detect, as shown in Equation \ref{eq:mAP}.

We have considered mAP metric because it provides a comprehensive measure of how well the model performs across all objects. In addition, we have considered P metric because it measures the accuracy of positive predictions made by the model and represents the the model's ability to make precise predictions of review texts.

\section{Results}

\begin{figure}
    \centering
    \includegraphics[width=1\linewidth]{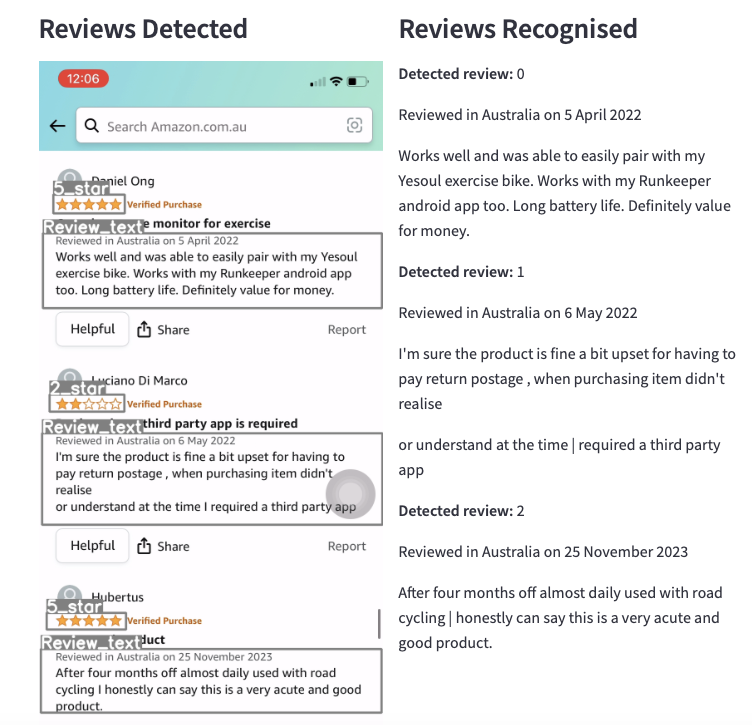}
    \caption{A sample review detection and recognition example}
    \label{fig:sample_result}
\end{figure}

In this section, we discuss the results and different applications of our proposed approach. At first, in Fig \ref{fig:sample_result} we show a sample result of our proposed approach. On the left side of Fig \ref{fig:sample_result}, ratings and the review texts detected by our model from a sample Amazon review page is shown. On the right side of Fig \ref{fig:sample_result}, recognised texts from the detected reviews are shown. It can be seen that, our model's detection accuracy as well as the recognition of detected review texts is also highly accurate. In addition, the recognised review texts are effectively grouped and there is a clear separation between the adjacent review texts.

The above example in Fig \ref{fig:sample_result} used a test image (i.e., not used in training) from the Amazon reviews platform. Our training data for the YOLOv8 model contains the samples from the same platform. Therefore, our proposed approach's high accuracy is expected. To investigate the robustness of our proposed approach, we tested some sample review images from different platforms for which our model is not trained. For that purpose, we have considered the following platforms: Menulog, Booking.com, Product Review, ebay. Although there are many review platforms exist and demonstrating results from every review platform will not be feasible in this paper. Therefore, we have selected only four test platforms. In the rest of this paper, each of these four platforms will be referred to as unknown test platform (UTP). Out of these four platforms, Menulog and Product Review allows to enter a star rating as well as a comment to the reviews. In contrast, Booking.com and ebay only allows to enter comments to the reviews.

We have evaluated on our trained YOLOv8 model on test database which consists of test images from the known platforms as well as four UTPs. The details of the test database is provided in Table \ref{tab:train_data}. The evaluation metrics we considered here is mAP for the overall model performance and P for the review text detection performance as explained in Section 5.1. The evaluation results are provided in table \ref{tab:results}.

\begin{table}
\centering
\caption{Results Summary}
\label{tab:results}
\begin{tabular}{p{2.5cm}||p{2cm}|p{2cm}|r}
\toprule
Source Website         &mAP (for overall detection) &Precision (for review text detection)  \\
\midrule
Amazon         &0.98    &  0.99   \\
Apple App Store        &0.97    & 0.99     \\  
Menulog           &0.87    &  0.89   \\ 
Product Review         &0.85    &  0.88   \\
Booking.com            &0.78    &  0.72  \\ 
eBay           &0.81    &   0.90 \\ 
\bottomrule
\end{tabular}
\end{table}

\begin{figure}
    \centering
    \includegraphics[width=1\linewidth]{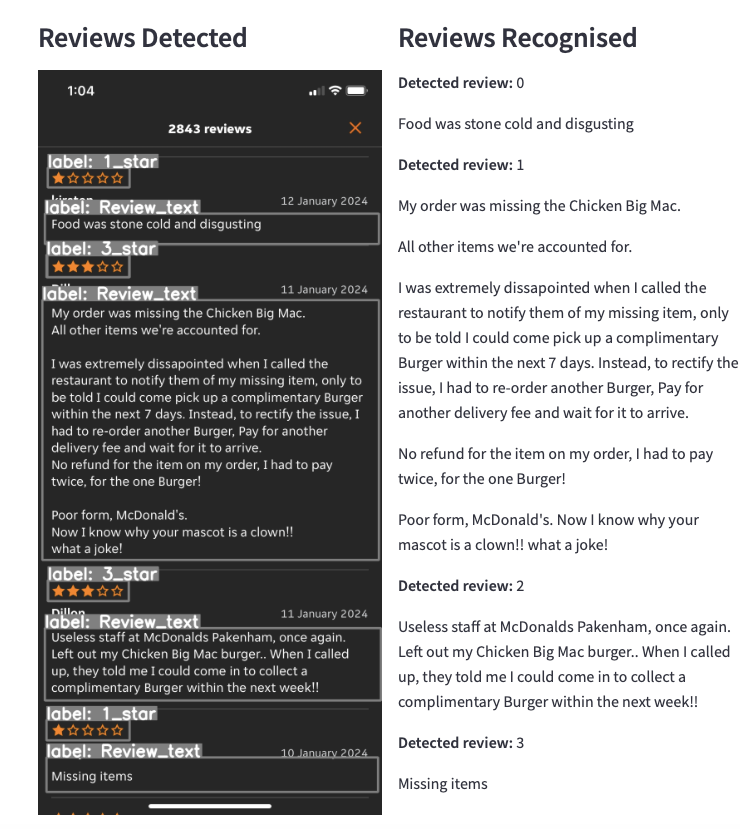}
    \caption{Review detection and recognition from Menulog}
    \label{fig:menulog_result}
\end{figure}

\begin{figure}
    \centering
    \includegraphics[width=1\linewidth]{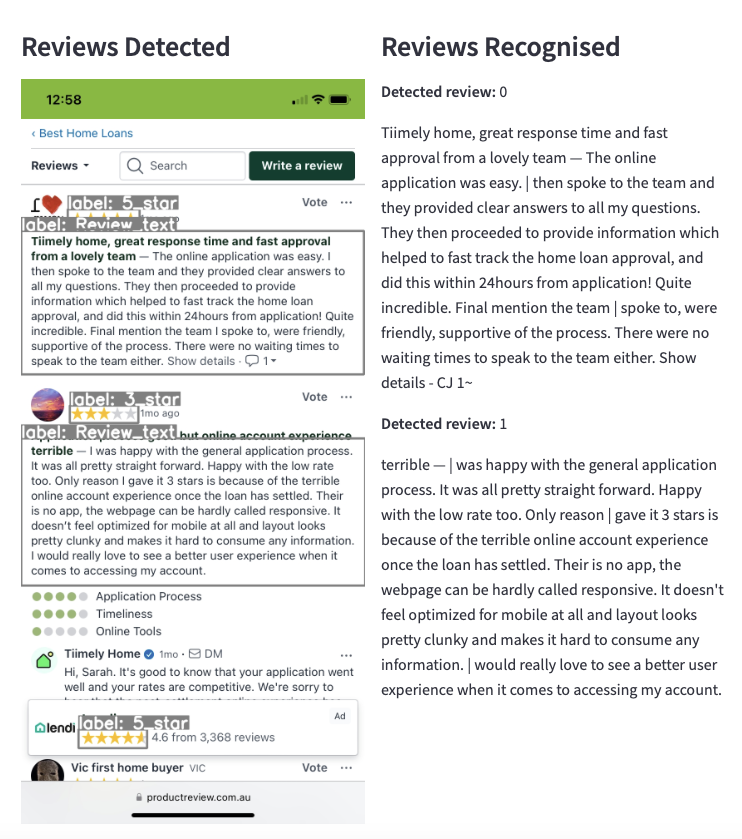}
    \caption{Review detection and recognition from Product Review}
    \label{fig:product_review_resul}
\end{figure}

\begin{figure}
    \centering
    \includegraphics[width=1\linewidth]{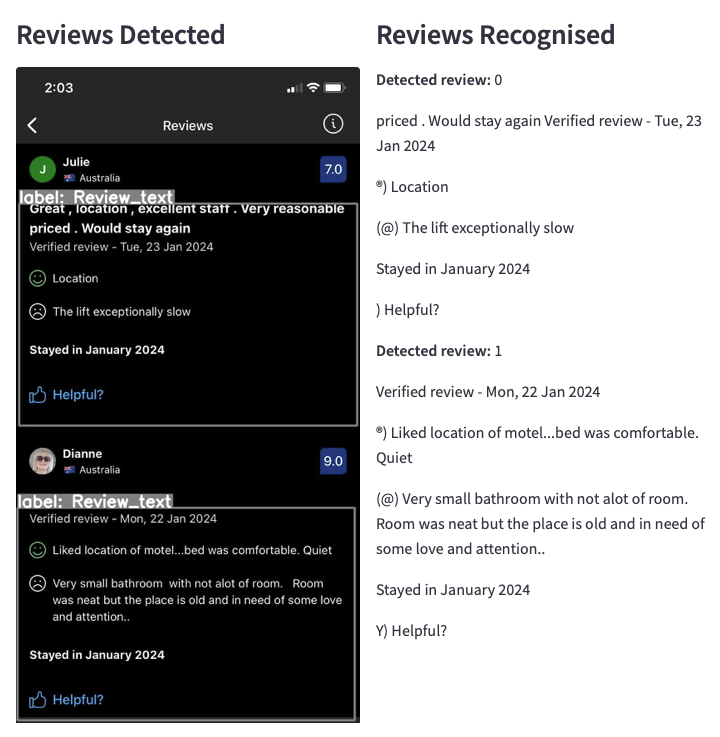}
    \caption{Review detection and recognition from Booking.com}
    \label{fig:booking_result}
\end{figure}

\begin{figure}
    \centering
    \includegraphics[width=1\linewidth]{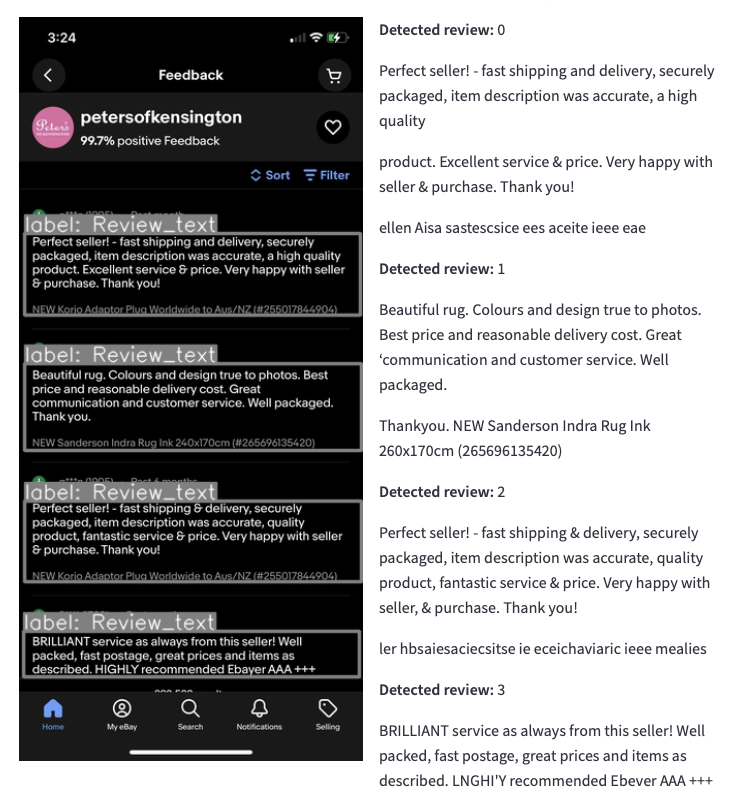}
    \caption{Review detection and recognition from ebay - I}
    \label{fig:ebay_result}
\end{figure}

\begin{figure}
    \centering
    \includegraphics[width=1\linewidth]{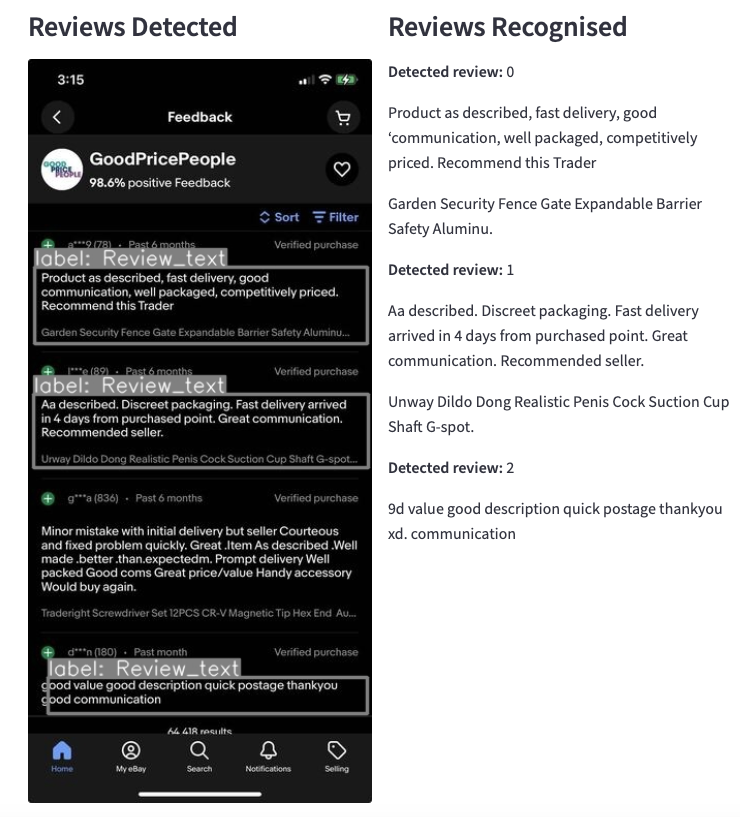}
    \caption{Review detection and recognition from ebay - II}
    \label{fig:ebay_result2}
\end{figure}

We have observed that our proposed framework is successfully able to detect and recognise reviews. Given the test images, our proposed framework can correctly detect the review texts as well as the star ratings in most cases as can be seen from the sample results on the four UTPs given by Figures \ref{fig:menulog_result}, \ref{fig:product_review_resul}, \ref{fig:booking_result}, \ref{fig:ebay_result}. In some cases, reviews remain undetected on test images from UTPs as can be seen by Figure \ref{fig:ebay_result2}. This can also be observed from the mAP scores obtained on UTP test images. However, the false positive detection rate of review texts specifically in the majority of UTPs is significantly lower, evident from the Precision values presented in Table \ref{tab:results}. Therefore, the information derived from the bulk processing of review texts on UTPs is reliable and trustworthy and resulting in effective decision-making processes in UTPs.


 A potential reason for the drop in performance when looking at domains outside the training data is that the test images from UTPs may differ slightly from those used for training. In addition, biases in the training dataset, specific to the platforms used, can lead to decreased performance when the model encounters new scenarios during testing on similar platforms \cite{navigli2023biases}. This is perfectly illustrated when we look at the screen shots for booking.com where our model performed the worst in the independent test. The review text contains indentation and custom icons that are specific to this website. In addition there is only a single numeric rating rather than the standard sequence of stars shown on other sites.


\section{Analysis Module}

The automated online review detection and recognition approach that we have proposed in this paper can be extended to many different applications by adding a task-specific module at the end. In this section, we outline three example applications: sentiment inconsistency analysis, multi-language support and fake review detection. 

Each of these applications can be considered a form of veracity filtering. Reviews that do not actually review the target product or service, or contain errors of inconsistencies in any way, should be removed from a website. These may be the result of user/system mistakes (writing about the wrong product or being allocated to the wrong product ID). Furthermore, some classes of undesirable behaviour may not be classified as fraudulent, such as posting irrelevant or spam content, but are nevertheless not genuine reviews. The broader task of determining whether a review is a genuine discussion of the product or not can be considered a form of veracity filter. If we are doing any kind of market analysis on the basis of reviews (such as sentiment analysis) then we want to pre-filter the data so that we look only at those reviews that genuinely discuss the product. 

\subsection{Sentiment Inconsistency Analysis}
One of the challenges for bulk processing of online reviews is filtering non-reliable reviews effectively so that no reliable reviews are filtered out \cite{zhang2023makes}. One of the approaches to find non-reliable reviews is to do sentiment inconsistent analysis (SIA). With the help of computer vision-based reviews detection and recognition, SIA can be performed easily for the review platforms that allow users to enter star ratings along with the comments. The individual reviews where the sentiments of the rating and the comment contradicts each other may mislead consumers and incorporating these type of reviews in bulk processing will have adverse effect on the overall understanding from the review insights. The reviews with inconsistent sentiments are also an indicator for it to be potentially a fake one.  

\begin{figure}
    \centering
    \includegraphics[width=1\linewidth]{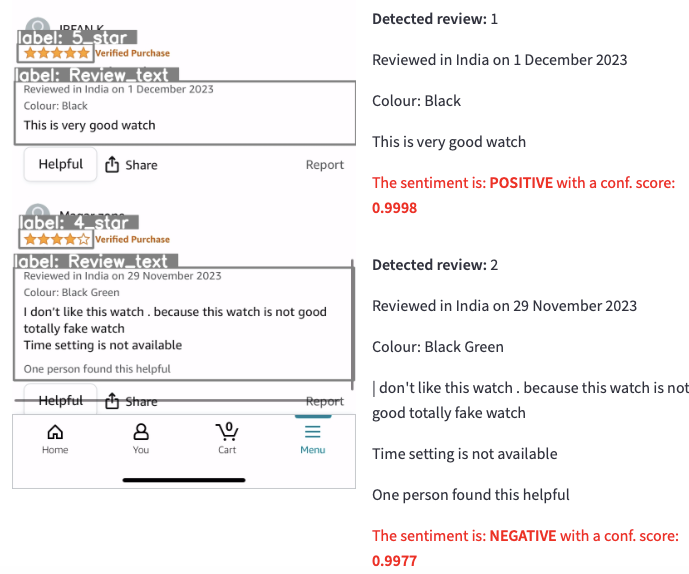}
    \caption{Illustration of Sentiment Inconsistency Analysis}
    \label{fig:sia}
\end{figure}

To perform SIA, we only added a sentiment analyser at the end of the text recognition in our proposed pipeline. For the sentiment analysis, we have used the Hugging Face transformer library \cite{wolf2019huggingface}. An example how we performed SIA is provided in Figure \ref{fig:sia} that shows a section of a review page with two reviews. For Detected review: 1, the sentiment of the rating (5 star can be interpreted as Positive) and the sentiment of the comment (Positive) are consistent. However, Detected review: 2, the sentiment of the rating (4 star can be interpreted as Positive) and the sentiment of the comment (Negative) are not consistent. In this way, any review that shows sentiment inconsistency like the Detected revirew: 2 in Figure \ref{fig:sia} can be easily detected with our proposed approach and then can be filtered out.

\subsection{Multi-language support}


Extracting reviews written in multiple languages from a review platform using HTML scraping presents significant challenges \cite{alzate2022mining}. Not only do many platforms prohibit HTML scraping, but the HTML codes themselves may lack uniformity to accommodate reviews in different languages. This lack of uniformity makes it difficult to devise a universal syntax for retrieving multilingual reviews. Additionally, even if a scraping code is meticulously designed, minor alterations made by platform developers could easily go unnoticed, resulting in the failure to collect a substantial amount of data.

\begin{figure}
    \centering
    \includegraphics[width=1\linewidth]{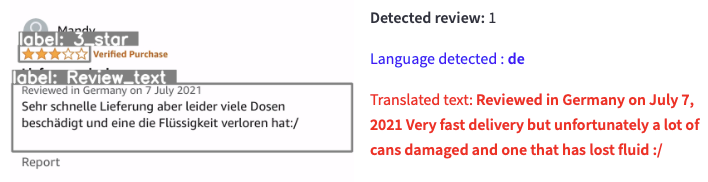}
    \caption{Illustration of review language detection and translation}
    \label{fig:multi_lang_support}
\end{figure}

All of the above issues can be addressed by our proposed approach, which combines review detection and text recognition, effectively tackles the aforementioned challenges. This is because, review text area detection is language independent and our OCR library supports most languages.  Furthermore, upon completing the text recognition process, we integrated a language translation module. This module first identifies the language of the review and then translates it into English as shown in Figure \ref{fig:multi_lang_support} that shows a section of the review page. This ensures that all gathered reviews are in English, streamlining the bulk processing of the reviews. The language translation module consists of langdetect \cite{nakatani2010langdetect} and googletrans \cite{googletrans} python libraries. 

\subsection{Fake review detection}
In this paper, we do not aim to conduct a research on fake online reviews detection. We only aim to demonstrate the flexibility of our proposed detection and recognition pipeline that can be directly used without any modification as a component in the fake reviews detection system. In other words, to detect fake reviews, at the end of our proposed pipeline, we incorporate an additional component that consist of a trained NLP model (trained to detect fake reviews).

\begin{figure}
    \centering
    \includegraphics[width=1\linewidth]{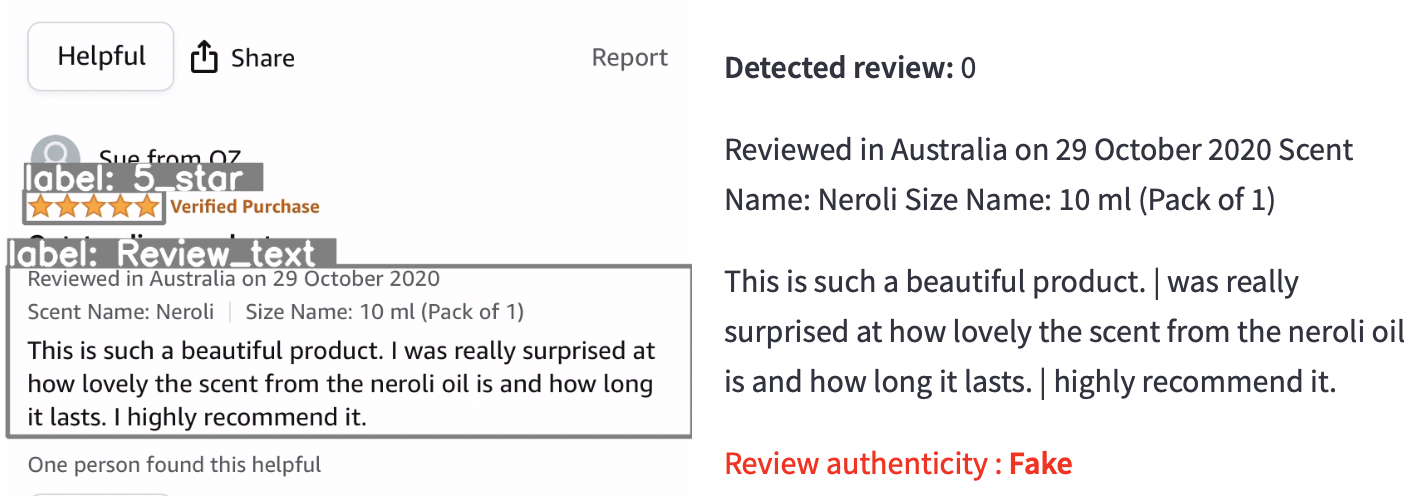}
    \caption{Fake review detection}
    \label{fig:review_auth}
\end{figure}

In this paper, the fake review detection is performed using ktrain python package \cite{maiya2022ktrain} which is a lightweight wrapper of Keras with the help of a large language model (LLM) of 7B-parameter provided by OnPrem python package \cite{onprem}.  The LLM was prompted to provide a decision whether a given review is fake or genuine. A sample of a detected fake review from a section of a review page is shown in Figure \ref{fig:review_auth}. In a real world application we would require a dataset of real and fake reviews to evaluate whether an LLM based prediction is reliable.

\section{Conclusion}

We have shown that using data from several online review websites we can build a computer vision machine learning system to detect reviews not only on known platforms but also on unknown ones. The method exploits the visual homogeneity of online review listings, and thereby is less susceptible to the outages that happen with HTML tag based filtering systems that are commonly used. We have illustrated that such a system can be extended into multiple analysis pipelines for filtering and reporting on the properties of online reviews. 

We observed that the review detection performance on the unknown test platforms is comparatively poorer than on those present in the training data. This issue can be solved by including samples from UTPs in the training data or by fine-tuning the base model with samples from UTPs. However, both of these approaches are time-consuming and it will violate our stance on the platform independence review detection. In our future work, we aim to solve this issue by integrating techniques such as knowledge distillation, few-shot learning, and zero-shot learning, so that the model can effectively mitigate the challenges posed by domain shift and dataset bias. Knowledge distillation enables the transfer of knowledge from a teacher model to a more compact student model, aiding in generalisation across different platforms. Few-shot learning empowers the model to learn from limited annotated examples, enabling adaptation to new platform characteristics with minimal data. Additionally, zero-shot learning leverages semantic information to recognise features unseen during training, further improving the model's adaptability to diverse platforms.

Despite potential for improvement, our modelling approach has delivered high review detection performance on multiple unknown platforms. In three of the four platforms the precision of review detection was approximately 90 percent, making it satisfactory for many applications. In addition, we have demonstrated multiple applications of our method for analysing reviews.

\balance

\bibliographystyle{IEEEtran}
\bibliography{reference}	
	
\end{document}